\documentclass[runningheads]{llncs}
\usepackage[T1]{fontenc}
\usepackage{graphicx,verbatim}
\usepackage{amsmath}
 \usepackage{booktabs} \usepackage{multirow}
\usepackage{amssymb}
\usepackage{bbm}

\begin{document}
\title{JANUS: Anatomy-Conditioned Gating for Robust CT Triage Under Distribution Shift}
%

\author{Lavsen Dahal\inst{1,2}\orcidID{0000-0002-8991-759X} \and
Yubraj Bhandari\inst{1,3}\orcidID{0009-0004-7279-4097} \and
Geoffrey Rubin\inst{4}\orcidID{0000-0002-3820-2500} \and 
Joseph Y. Lo\inst{1,2}\orcidID{0000-0002-9540-5072}}
\authorrunning{Dahal et al.}
%
\institute{Center for Virtual Imaging Trials, RAI Labs, Department of Radiology, Duke University, Durham NC 27708, USA \and
Electrical and Computer Engineering, Pratt School of Engineering, Duke University, Durham, NC 27708, USA
\and
Department of Mathematics, Trinity College of Arts \& Sciences, Duke University, Durham, NC 27708, USA
\and
Department of Radiology and Imaging Sciences, University of Arizona College of Medicine, Tucson, AZ 85004, USA\\
\email{\{lavsen.dahal,joseph.lo\}@duke.edu}}

  
\maketitle              

\begin{abstract}
Automated CT triage requires models that are simultaneously accurate 
across diverse pathologies and reliable under institutional shift. 
While Vision Transformers provide strong visual representations, 
many clinically significant findings are defined by quantitative imaging biomarkers rather than appearance alone. We introduce \textbf{JANUS}, a physiology-guided dual-stream architecture that conditions visual embeddings on macro-radiomic priors via \emph{Anatomically Guided Gating}. On the MERLIN test set ($N{=}5{,}082$), JANUS attains macro-AUROC $0.88$ and AUPRC $0.74$, outperforming all reproduced baselines. It generalizes to an external dataset ($N{=}2{,}000$; AUROC $0.87$), with the largest gains on findings defined by size and attenuation as well as improved calibration on both datasets.  We further quantify prediction suppression using the \emph{Physiological Veto Rate} (PVR), showing that under domain shift JANUS reduces high-confidence false positives substantially more often than true positives. Together, these results are consistent with physically grounded conditioning that improves both discrimination and reliability in CT triage. Code is made publicly available at github repository \url{https://github.com/lavsendahal/janus} and model weights are at \url{https://huggingface.co/lavsendahal/janus} .
\end{abstract}


\section{Introduction}
Automated CT triage seeks to flag clinically significant findings, 
many defined by quantitative thresholds rather than appearance alone: 
organomegaly by size, aneurysm by vessel caliber, calcific burden 
by HU density~\cite{linguraru2012assessing,gucuk2014usefulness,radiopaedia_splenomegaly,paudyal2023artificial}. Missing such 
measurement-defined findings can have major consequences, e.g., an 
undetected aneurysm can rapidly escalate to a life-threatening 
event. As CT utilization grows and radiologist workloads 
intensify, reliable automated triage becomes a clinical 
necessity~\cite{momin2025systematic,winder2021we}.

CT foundation models have emerged to meet this need, learning rich 
visual representations via large-scale 
pretraining~\cite{pai2025vision,momin2025systematic,blankemeier2024merlin,shui2025large}, yet lack explicit mechanisms for the measurement 
cues that define many urgent findings~\cite{li2023transforming}. 
Under distribution shift, this gap becomes consequential: underlying 
anatomy is protocol-invariant, but visual appearance can shift with 
scanner and reconstruction settings~\cite{guo2024impact}.

Prior attempts to bridge this gap incorporated only coarse 
organ-level statistics (mean volume and mean HU) as auxiliary 
inputs to visual embeddings~\cite{dahal2026organ}, demonstrating that quantitative priors 
carry signal beyond visual features. \emph{Macro-Radiomics}~\cite{dahal2026ctidpsegmentationderivedquantitativephenotypes} enables 
a richer, clinically aligned measurement space, including organ-level descriptors, diameters, 
densitometry profiles, calcific burden, and fluid volumes. Unlike microscopic texture features, Macro-radiomics operate 
at the scale of radiological decision-making and are reliably 
computable for large, well-defined organs via modern segmentation 
models~\cite{wasserthal2023totalsegmentator,dahal2025xcat}.

Yet how to enforce such measurements as structural constraints remains unaddressed. Additive fusion lacks 
an explicit suppression mechanism: under distribution shift, visual 
models could latch onto spurious correlations~\cite{geirhos2020shortcut} 
and concatenation may not encourage selective suppression when predictions conflict with available quantitative priors. Multiplicative 
gating~\cite{perez2018film,hu2018squeeze} provides this inductive 
bias: physical measurements modulate visual evidence through a 
learned bottleneck, structurally enabling suppression rather than 
merely shifting predictions.

We introduce \textbf{JANUS}, a physiology-guided architecture 
that operationalizes this principle: quantitative measurements should 
constrain visual predictions, not merely augment them.

\noindent\textbf{Contributions.}
\begin{itemize}
\item \textbf{Architecture.} JANUS integrates the 
full macro-radiomic phenotype space with a visual stream 
via \emph{Anatomically Guided Gating}: a disease-specific 
multiplicative bottleneck that modulates visual evidence 
through scalar priors, providing an explicit 
suppression mechanism.

\item \textbf{Metric.} We introduce the \emph{Physiological Veto Rate} (PVR) to quantify suppression of high-confidence baseline false positives at fixed thresholds. To assess whether suppression is selective rather than uniform, we additionally report \emph{veto selectivity}, defined as the ratio of false-positive suppression to true-positive suppression. On the external dataset, veto selectivity reaches $10.8\times$, indicating suppression is strongly concentrated on high-confidence false positives while largely preserving true positives.

\item \textbf{Empirical.} JANUS achieves the highest macro-AUROC among reproduced baselines on MERLIN and maintains performance on an external dataset, while improving calibration (lower ECE) on both; results support macro-radiomic priors improving discrimination and reliability.

\end{itemize}


\section{Methods}

\begin{figure*}[t]
    \centering
    \includegraphics[width=\textwidth]{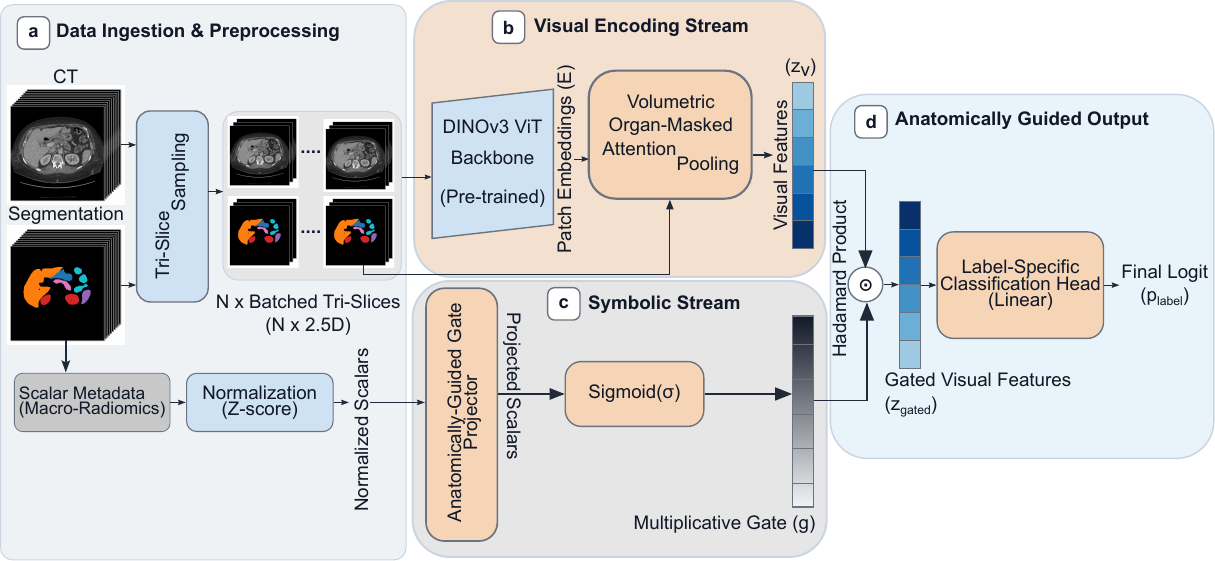}
\caption{\textbf{JANUS Architecture.} 
\textbf{(a)} A 3D CT volume is sampled into $N$ 2.5D tri-slices; 
segmentation masks yield macro-radiomic scalar priors. 
\textbf{(b)} A DINOv3 backbone extracts patch embeddings condensed 
into a label-specific visual feature $z_v$ via Organ-Masked Attention 
Pooling. 
\textbf{(c)} Scalar priors are projected and sigmoid-bounded to form 
a physiological gate $g$. 
\textbf{(d)} $g$ modulates $z_v$ via Hadamard product ($\odot$), 
acting as a \emph{Physiological Veto}; gated features $z_{\text{gated}}$ 
are passed to a label-specific linear head.}
    \label{fig:janus_architecture}
\end{figure*}

\subsection{Problem Formulation and the Geometric Gap}
We consider automated triage as a multi-label classification problem over a CT volume
$X\in\mathbb{R}^{D\times H\times W}$ with partially observed targets
$\mathbf{y}\in\{0,1,\emptyset\}^L$, where $\emptyset$ denotes missing labels.
Let $\delta_\ell=\mathbbm{1}[y_\ell\neq\emptyset]$ indicate label availability.
Our goal is to learn a physiologically conditioned predictor
$P(\mathbf{y}\mid X,S)$, where $S$ encodes segmentation-derived macro-radiomic priors
that provide explicit \emph{quantitative measurement context}
spanning anatomical scale and tissue densitometry.

We initialize the visual backbone with DINOv3 
weights~\cite{simeoni2025dinov3}. Standard attention 
pooling is invariant to token count by construction: 
geometric quantities such as organ volume or vessel 
diameter are encoded in how many tokens an organ 
occupies, not per-token appearance, and is thus not explicitly 
represented in the pooled embedding, the \emph{pooling-induced 
geometric gap}. We address this by conditioning visual 
evidence on explicit scalar priors $S$ derived from 
segmentation.

\subsection{JANUS Architecture}
We introduce \textbf{JANUS} (Fig.~\ref{fig:janus_architecture}), a dual-stream
physiology-guided model that couples (i) a visual stream extracting appearance features
from $X$ and (ii) a symbolic stream encoding macro-radiomic priors. JANUS fuses the two
streams through a multiplicative gate that modulates the visual embedding using
explicit anatomical measurements.

\noindent\textbf{Visual Stream: 2.5D Tri-slice Encoding.}
Given $X$, we construct a 2.5D representation by sampling axial slice centers
$\{t\}_{t=1}^{T}$ with stride $s$ and forming a 3-channel input at each center by
stacking adjacent slices:
$[X_{t-1},X_t,X_{t+1}]$ (boundary slices are replicated). Each tri-slice is resized to a
fixed resolution and normalized with ImageNet statistics. Passing all tri-slices through
a DINOv3 ViT produces patch token embeddings. We discard the CLS token and DINOv3 register
tokens and reshape tokens as $\{u_i\}_{i=1}^{N}$ per tri-slice, yielding
$u\in\mathbb{R}^{B\times T\times N\times d}$, where $B$ is the 
batch size, $N$ is the number of spatial patch tokens per tri-slice, and 
$d$ is the embedding dimension ($d{=}768$ for ViT-B).

\noindent\textbf{Dilated ROI-Masked Attention Pooling.}
For each label $\ell$, we define a disease-specific 3D ROI mask $M_\ell$ 
by composing segmentation channels relevant to that pathology: a single 
organ for focal findings (e.g., aorta for Abdominal Aortic Aneurysm), a multi-organ union 
where appropriate (e.g., liver and spleen for hepatic steatosis), or 
a localization box where no organ proxy exists (e.g., appendicitis).
To capture peri-organ context (e.g., fat stranding), we apply a resolution-adaptive
morphological dilation of $M_\ell$ by a physical radius $r_\ell$ (mm), using voxel spacing
to convert millimeters to per-axis kernel sizes. We align the 3D ROI mask with the tri-slice sampling by extracting a 2D mask at each tri-slice center $t$ and resizing it to the encoder input grid. We then downsample this ROI to the ViT
patch grid via nearest-neighbor interpolation to obtain a binary token mask
$m_{t,i,\ell}\in\{0,1\}$ indicating whether spatial position $i$ in tri-slice $t$ lies inside the (dilated) ROI.

Given tokens $\{u_{t,i}\}$ where $t\in\{1,\dots,T\}$ indexes tri-slices and 
$i\in\{1,\dots,N\}$ indexes spatial positions, JANUS pools a label-specific visual 
embedding $z_{v,\ell}$ using ROI-masked attention:
\begin{equation}
\begin{aligned}
a_{t,i,\ell}&=\frac{\phi_\ell(u_{t,i})+\beta_{\text{in},\ell}\, m_{t,i,\ell}}{\tau_\ell},\quad
w_{t,i,\ell}=\mathrm{softmax}_{i:\,m_{t,i,\ell}=1}\!\left(a_{t,i,\ell}\right),\\[6pt]
z_{v,\ell}&=\frac{1}{T}\sum_{t=1}^{T}\sum_{i:\,m_{t,i,\ell}=1} w_{t,i,\ell}\,u_{t,i}.
\end{aligned}
\end{equation}
Here $a_{t,i,\ell}$ denotes the unnormalized attention logit for token $i$ at tri-slice $t$
(label $\ell$) prior to softmax normalization; $\phi_\ell(\cdot)$ is a learned linear scorer,
$\beta_{\text{in},\ell}$ is a learnable logit bias for ROI tokens, and $\tau_\ell$ is a
learnable temperature. The softmax normalizes over spatial positions $i$ within the ROI
\emph{independently per slice} $t$, so $\sum_{i:\,m_{t,i,\ell}=1} w_{t,i,\ell}=1$ for each $t$.
Outside tokens receive zero weight; if an ROI is empty for a sample, we fall back to
uniform weights over all tokens.

\noindent\textbf{Symbolic Stream: Anatomically Guided Gating.}
From the same anatomical masks used to define $M_\ell$, we compute 
a disease-specific macro-radiomic prior vector $s_\ell\in\mathbb{R}^{K_\ell}$. Each feature is z-score normalized using training-set statistics.

We map the prior vector to a physiological gate $g_\ell\in[0,1]^d$ and modulate
visual evidence by element-wise multiplication:
\begin{equation}
g_\ell=\sigma(W_{g,\ell}s_\ell+b_{g,\ell}),
\qquad
z_{\text{gated},\ell}= z_{v,\ell}\odot g_\ell,
\end{equation}
where $W_{g,\ell}\in\mathbb{R}^{d\times K_\ell}$ and $b_{g,\ell}\in\mathbb{R}^{d}$ 
are learned. Because $g_\ell\in[0,1]^d$, the gate implements a bounded, dimension-wise
re-weighting of the visual embedding conditioned on quantitative measurements, which can
attenuate components of the representation when the priors provide countervailing evidence.
We initialize $b_{g,\ell}=2.0$ so that $g_\ell\approx0.88$ at 
initialization, keeping the gate mostly open and ensuring stable 
gradient flow. This initialization biases the model toward preserving the visual stream
unless training data support stronger gating, which can mitigate failure modes when scalar
priors are weakly informative.

\subsection{Prediction and Training Objective}
For each label $\ell$, we apply a disease-specific linear head to obtain logits
$\hat{y}_\ell=\mathbf{w}_\ell^\top z_{\text{gated},\ell}+b_\ell$ and probabilities
$p_\ell=\sigma(\hat{y}_\ell)$. We optimize a curriculum-weighted BCE over 
partially observed labels $y_\ell\in\{0,1,\emptyset\}$:
\begin{equation}
\mathcal{L}^{(e)}=\frac{
\sum_{\ell=1}^{L}\bigl[\delta_\ell\cdot\mathrm{BCE}(p_\ell,y_\ell)
+(1-\delta_\ell)\cdot w^{(e)}\cdot\mathrm{BCE}(p_\ell,0)\bigr]
}{\sum_{\ell=1}^{L}\bigl[\delta_\ell+(1-\delta_\ell)\cdot w^{(e)}\bigr]}
\end{equation}
where $w^{(e)}$ ramps linearly from $0$ to $w_{\max}=0.3$ over epochs 
$E_{\text{ignore}}=10$ to $E_{\text{ignore}}+E_{\text{ramp}}=20$, 
treating missing labels as weak negatives thereafter.

\section{Experimental Setup}

\noindent\textbf{Datasets.}
We train and evaluate on MERLIN~\cite{blankemeier2024merlin}, a retrospective 
abdominal CT dataset ($N{=}25{,}275$; 30 labels; splits 15,175/5,018/5,082). 
We additionally evaluate on an external dataset from a different [redacted] U.S.\ hospital 
($N{=}2{,}000$), with labels derived from radiology reports via dual-LLM 
consensus: a finding is positive only if both Qwen-3~\cite{yang2025qwen3,agrawal2025pillar} 
and MedGemma~\cite{sellergren2025medgemma} independently agree, and negative 
only if both confirm absence; abstaining studies are excluded, with explicit 
contradictions in fewer than $0.5\%$ of cases. Organ masks are obtained via 
TotalSegmentator~\cite{wasserthal2023totalsegmentator}.

\noindent\textbf{Baselines.} We compare against three baselines: (i)~\textit{ViT-Baseline}, DINOv3 ViT-B/16 with global average pooling and no scalar integration; (ii)~\textit{ORACLE-CT}~\cite{dahal2026organ}, a recent concurrent anatomy-aware baseline; and (iii)~\textit{ORACLE-CT+OSF}~\cite{dahal2026organ}, which reported the strongest supervised results on the official MERLIN test split and injects scalar priors via projection+concatenation at the pooled embedding (originally mean volume/HU). In our experiments, ORACLE-CT+OSF uses the same macro-radiomic prior bank and backbone as JANUS, isolating prior usage (projection+concatenation vs.\ prior-modulated features).

\noindent\textbf{Training.}
Volumes are clipped to $[-1000,1000]$~HU, resampled to 
$1.5{\times}1.5{\times}3.0$~mm, and center-cropped to 
$224{\times}224{\times}160$, tri-slice stride (s = 1). Augmentations include 3D affine transforms, gamma jitter, 
and noise. DINOv3 
ViT-B/16 ~\cite{simeoni2025dinov3} is trained 
end-to-end for 20 epochs on 4$\times$A6000 GPUs (batch 
size 10).

\noindent\textbf{Evaluation.}
All AUROC and AUPRC values are macro-averaged over labels.
Calibration is measured via ECE (10 equal-width bins)~\cite{guo2017calibration} .

\noindent\textbf{Physiological Veto Rate (PVR).}
For label $\ell$, let $\mathcal{F}_\ell=\{i:\; y_{i\ell}=0,\; p^{\mathrm{ViT}}_{i\ell}\ge 0.8\}$ denote high-confidence false positives under the ViT baseline. PVR measures the fraction of these baseline errors suppressed by JANUS below a decision threshold:
\begin{equation}
\mathrm{PVR}_\ell
=\frac{1}{|\mathcal{F}_\ell|}
\sum_{i\in\mathcal{F}_\ell}\mathbb{1}\!\left[p^{\mathrm{JANUS}}_{i\ell}<0.5\right].
\end{equation}
We report the mean over labels with $|\mathcal{F}_\ell|\ge 5$. We additionally report the true-positive suppression rate $\mathrm{TSR}_\ell=\Pr(p^{\mathrm{JANUS}}_{i\ell}<0.5\mid y_{i\ell}=1)$ and summarize \emph{veto selectivity} as $\mathrm{PVR}_\ell/\mathrm{TSR}_\ell$, where values $\gg 1$ indicate suppression concentrated on baseline false positives rather than uniformly reducing confidence.

\section{Results}

\subsection{Global Performance, Reliability, and Robustness}

\begin{table}[tb]
\centering
\caption{\textbf{Global benchmarking.} AUROC, AUPRC, and ECE (95\% bootstrap CIs; $n{=}1000$ resamples) on MERLIN (internal) and Duke-Abdomen (external) datasets (27 shared labels). \textbf{JANUS} achieves the strongest overall performance across metrics.}
\label{tab:global}
{%
\fontsize{8pt}{9pt}\selectfont  
\begin{tabular}{@{} l ccc ccc @{}}
\toprule
& \multicolumn{3}{c}{MERLIN (Internal, $N=5{,}082$)} 
& \multicolumn{3}{c}{Duke-Abdomen (External, $N=2{,}000$)} \\
\cmidrule(lr){2-4} \cmidrule(lr){5-7}
\textbf{Model} &
AUC $\uparrow$ & AUPRC $\uparrow$ & ECE $\downarrow$ &
AUC $\uparrow$ & AUPRC $\uparrow$ & ECE $\downarrow$ \\
\midrule
ViT-Baseline~\cite{simeoni2025dinov3}
& 0.84 & 0.66 & 0.13 & 0.84 & 0.67 & 0.16 \\
& (0.83--0.85) & (0.64--0.68) & (0.12--0.14)
& (0.82--0.85) & (0.65--0.70) & (0.16--0.17) \\
ORACLE-CT~\cite{dahal2026organ}
& 0.86 & 0.69 & 0.15 & 0.85 & 0.69 & 0.19 \\
& (0.85--0.87) & (0.68--0.72) & (0.14--0.16)
& (0.83--0.86) & (0.67--0.72) & (0.18--0.20) \\
ORACLE-CT+OSF~\cite{dahal2026organ}
& 0.84 & 0.66 & 0.17 & 0.81 & 0.66 & 0.22 \\
& (0.83--0.85) & (0.65--0.68) & (0.16--0.18)
& (0.80--0.83) & (0.64--0.68) & (0.21--0.23) \\
\textbf{JANUS (Ours)}
& \textbf{0.88} & \textbf{0.74} & \textbf{0.09}
& \textbf{0.87} & \textbf{0.72} & \textbf{0.15} \\
& (0.87--0.89) & (0.72--0.76) & (0.09--0.11)
& (0.85--0.88) & (0.71--0.75) & (0.15--0.16) \\
\bottomrule
\end{tabular}
}
\end{table}

\begin{figure*}[!t]
\centering
\includegraphics[width=1.0\textwidth]{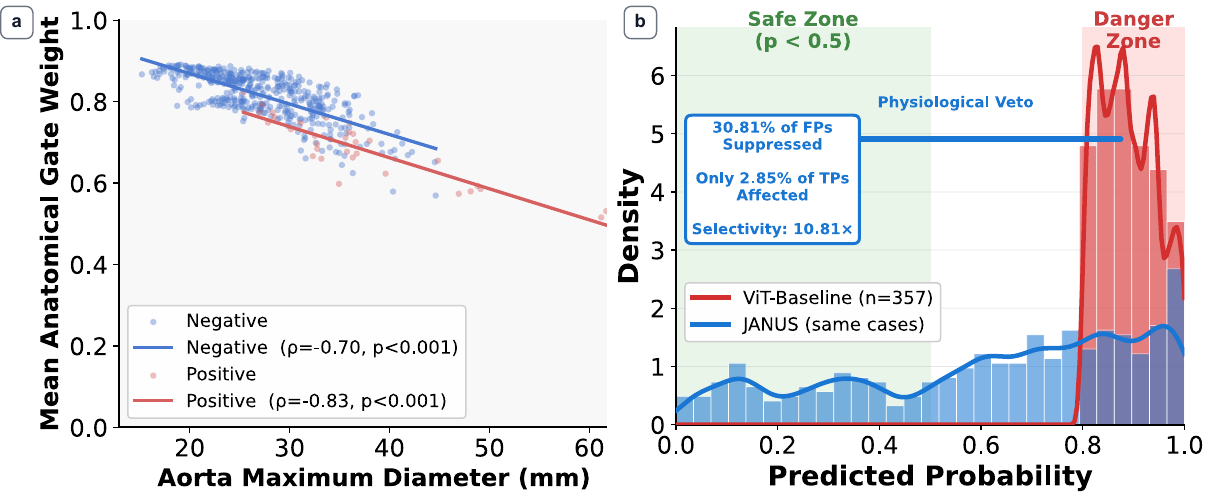}
\caption{\textbf{External Dataset: Gate behavior and physiological veto.}
\textbf{(a)} Mean anatomical gate weight (mean of $g=\sigma(Ws+b)\in(0,1)^d$) versus aortic maximum diameter for AAA negatives and positives; lower values indicate stronger down-weighting of visual features.
\textbf{(b)} PVR on the external dataset: among overconfident ViT false positives ($p_{\mathrm{ViT}}\!\ge\!0.8$, $n{=}357$), JANUS reduces $p<0.5$ more often for negatives than positives, consistent with selective suppression under shift.}
\label{fig:gate_and_physiology}
\end{figure*}

\begin{figure*}[!t]
\centering
\includegraphics[width=1.0\textwidth]{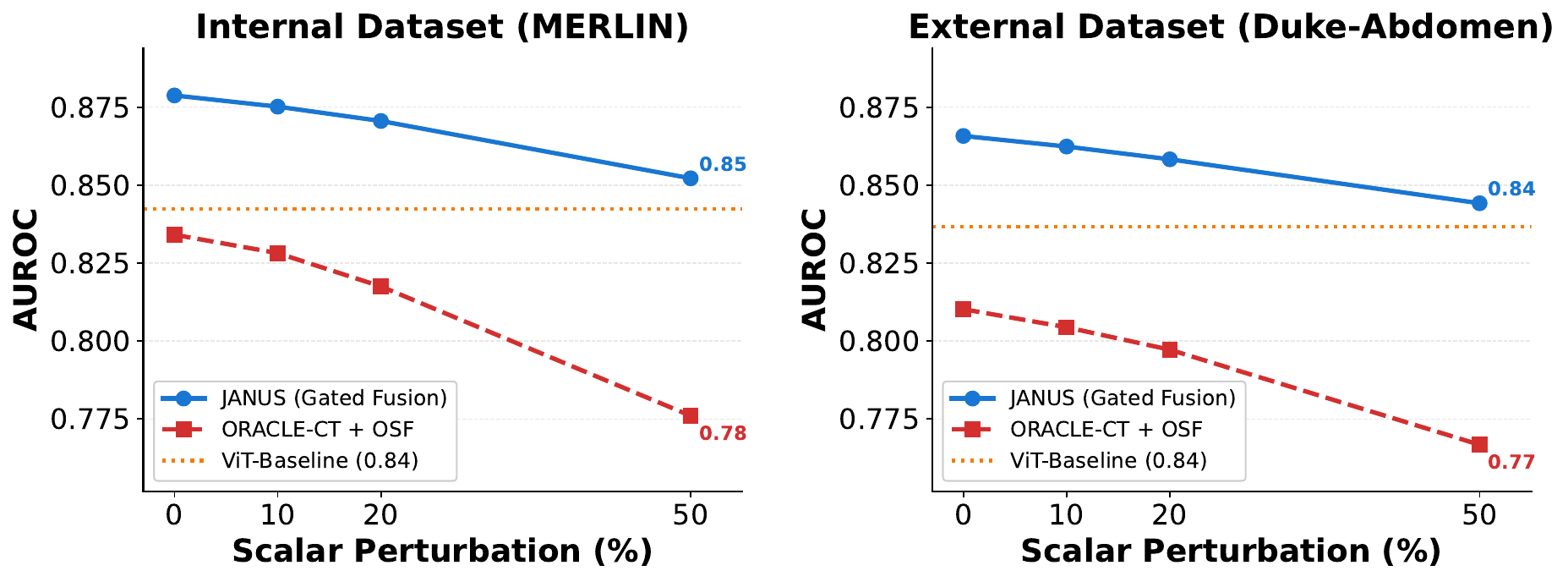}
\caption{\textbf{Robustness to scalar corruption.} AUROC under 
10\%, 20\%, and 50\% corruption of macro-radiomic priors on 
internal and external datasets. JANUS degrades gracefully and 
remains above the ViT-Baseline (dotted) at all corruption 
levels, including 50\%, suggesting the multiplicative gate 
bounds the influence of corrupted inputs.}
\label{fig:perturbation}
\end{figure*}

\begin{table*}[t]
\centering
\caption{\textbf{Stratified performance by pathology.} AUROC on 
MERLIN (internal) and external (Duke-Abdomen) datasets across 27 shared 
labels. Group rows report category means; two representative 
diseases are shown per group. \textbf{Gain} ($\Delta$AUROC, 
JANUS vs.\ ViT-Baseline, external) is largest where physiology 
scalars are well-defined (Geometric, Densitometric, Chronic). As expected,   degradation is minimal for Focal/Control Group, where scalar priors are inherently uninformative.}
\label{tab:stratified_results}

{%
\fontsize{8pt}{9.2pt}\selectfont
\setlength{\tabcolsep}{1.3pt}
\renewcommand{\arraystretch}{1}

\begin{tabular}{@{} l|cccc|cccc|c @{}}
\toprule
& \multicolumn{4}{c|}{\textbf{MERLIN}} & \multicolumn{4}{c|}{\textbf{EXTERNAL}} & \textbf{Gain} \\
\cline{2-5}\cline{6-9}
\textbf{Pathology} & ViT & ORACLE & +OSF & \textbf{JANUS} & ViT & ORACLE & +OSF & \textbf{JANUS} & (Ext.) \\
\midrule

\textbf{Geometric} (7)
& 0.89 & 0.92 & 0.90 & \textbf{0.94}
& 0.86 & 0.89 & 0.88 & \textbf{0.91} & +0.05\\
\quad Prostatomegaly
& 0.80 & 0.84 & 0.88 & 0.89
& 0.74 & 0.83 & 0.89 & 0.88 & +0.14 \\
\quad Aortic Aneurysm
& 0.83 & 0.91 & 0.94 & 0.98
& 0.88 & 0.87 & 0.95 & 0.96 & +0.08 \\
\midrule

\textbf{Densitometric} (5)
& 0.83 & 0.84 & 0.88 & \textbf{0.90}
& 0.82 & 0.83 & 0.89 & \textbf{0.88} & +0.06 \\
\quad Hepatic Steatosis
& 0.79 & 0.82 & 0.95 & 0.96
& 0.79 & 0.81 & 0.88 & 0.88 & +0.09 \\
\quad Gallstones
& 0.72 & 0.72 & 0.80 & 0.84
& 0.54 & 0.62 & 0.80 & 0.81 & +0.27 \\
\midrule

\textbf{Fluid/Global} (3)
& 0.97 & 0.97 & 0.93 & \textbf{0.97}
& 0.98 & 0.98 & 0.95 & \textbf{0.98} & 0.0 \\
\quad Pleural Effusion
& 0.97 & 0.97 & 0.93 & 0.97
& 0.98 & 0.98 & 0.94 & 0.98 & 0.0 \\
\quad Ascites
& 0.95 & 0.95 & 0.89 & 0.95
& 0.97 & 0.97 & 0.92 & 0.97 & 0.0 \\
\midrule

\textbf{Chronic} (6)
& 0.85 & 0.87 & 0.85 & \textbf{0.90}
& 0.82 & 0.83 & 0.83 & \textbf{0.87} & +0.05 \\
\quad Renal Cyst
& 0.81 & 0.82 & 0.82 & 0.89
& 0.81 & 0.75 & 0.87 & 0.95 & +0.14 \\
\quad Pancreatic Atrophy
& 0.80 & 0.87 & 0.90 & 0.90
& 0.72 & 0.84 & 0.81 & 0.83 & +0.11 \\
\midrule

\textbf{Focal/Control} (6)
& 0.73 & \textbf{0.75} & 0.65 & 0.73
& 0.76 & \textbf{0.75} & 0.57 & 0.74 & -0.02 \\
\quad Appendicitis
& 0.66 & 0.67 & 0.59 & 0.65
& 0.85 & 0.86 & 0.53 & 0.81 & -0.04 \\
\quad Fracture
& 0.78 & 0.80 & 0.68 & 0.75
& 0.80 & 0.84 & 0.75 & 0.78 & -0.02 \\
\bottomrule
\end{tabular}
}
\end{table*}

\noindent\textbf{Overall performance.}
Table~\ref{tab:global} reports global results; unless otherwise stated, $\Delta$AUROC is measured on the external dataset relative to ViT-Baseline. ORACLE-CT shows organ-level context adds signal ($\Delta$AUROC $= {+0.02}$). Additive scalar fusion (+OSF) improves size and density-driven findings but does not improve overall external macro-AUROC ($\Delta$AUROC $= {-0.03}$) and has the highest external ECE. JANUS better leverages macro-radiomic priors via anatomically guided gating, achieving the strongest performance on both datasets (AUROC/AUPRC $0.88$/$0.74$ internal, $0.87$/$0.72$ external) with the lowest ECE on both.

\noindent\textbf{Gate Mechanism and Physiological Veto.}
Figure~\ref{fig:gate_and_physiology}(a) examines how the learned gate varies with a quantitative biomarker on the external dataset. Using Abdominal Aortic Aneurysm (AAA) as a probe, the \emph{mean anatomical gate weight} (mean of $g=\sigma(Ws+b)\in(0,1)^d$) decreases monotonically with aortic maximum diameter within both label groups, consistent with sensitivity to the scalar measurement rather than simply mirroring disease status. Across all 27 pathologies (Fig.~\ref{fig:gate_and_physiology}b), among baseline overconfident false positives, JANUS reduces $30.8\%$ below threshold while affecting only $2.85\%$ of true positives ($10.8\times$ selectivity). Because PVR is computed per label, operating points could be set per pathology to trade off false-positive reduction against true-positive retention.

\noindent\textbf{Scalar robustness.} To stress-test 
sensitivity to scalar quality, we corrupt each feature 
by uniform random noise scaled to $p$\% of its value 
at inference ($p \in \{10, 20, 50\}$), spanning mild 
to deliberately extreme degradation. JANUS degrades 
more gracefully across all levels on both datasets, 
remaining above the ViT baseline even at $p{=}50$ 
(external AUROC $0.84$) while +OSF approaches it 
($0.77$), consistent with the sigmoid-bounded gate 
providing a structural bound on corrupted inputs 
(Figure~\ref{fig:perturbation}).

\subsection{Stratified Analysis by Pathology}
Table~\ref{tab:stratified_results} stratifies performance 
by diagnostic mechanism, grouping findings by the type of 
physical signal available to the gate.

\noindent\textbf{Measurement-defined targets.}
JANUS yields the largest gains on geometric and densitometric 
findings, where scalar measurements directly encode the 
diagnostic criterion (group means $\Delta$AUROC $= {+0.05}$ and 
$= {+0.06}$ externally). Fluid/Global findings saturate near 
$0.98$ AUROC across all models, leaving no headroom. Chronic 
targets show consistent improvement ($\Delta$AUROC $= {+0.05}$), 
with Renal Cyst benefiting most ($\Delta$AUROC $= {+0.14}$).

\noindent\textbf{Focal targets and non-local priors.}
Focal pathologies provide a natural stress test for macro-radiomic priors, which summarize organ-level or global measurements and are not designed to capture fine spatial localization. Accordingly, JANUS shows only a small change on these targets ($\Delta$AUROC $= {-0.02}$). In contrast, +OSF decreases substantially ($\Delta$AUROC $= {-0.19}$), suggesting that directly fusing non-local scalar features can yield an unfavorable accuracy.

\section{Conclusion}
We introduced \textbf{JANUS}, a physiology-guided framework that integrates 
a comprehensive macro-radiomic prior bank with organ-localized visual 
representations via disease-specific multiplicative gating.  Under a matched training and evaluation protocol using released baseline code, JANUS achieves the highest macro-AUROC on MERLIN ($0.88$) and generalizes under distribution shift to an external dataset (AUROC $0.87$), with substantially improved calibration on both datasets, suggesting an association between physical grounding and imporved accuracy and reliability.

\noindent\textbf{Limitations.}
JANUS relies on a sequential pipeline where segmentation precedes 
gating, limiting guidance for focal findings where organ-level priors 
provide limited signal. Scalar robustness is evaluated via simulated 
corruption rather than realistic segmentation errors, and evaluation 
with CT-pretrained backbones remains future work.

\section*{Acknowledgements}
This work was funded by the Center for Virtual Imaging Trials, NIH grants P41EB028744, R01EB001838, and R01CA261457.

\section*{Disclosure of Interests.} The authors have no competing interests to declare that are relevant to the content of this article.
\bibliographystyle{splncs04}
\bibliography{mybibliography}

\end{document}